\documentclass{article}

\PassOptionsToPackage{numbers, compress}{natbib}
    \usepackage[final]{neurips_attrib_2024}

\usepackage[utf8]{inputenc} % allow utf-8 input
\usepackage[T1]{fontenc}    % use 8-bit T1 fonts
\usepackage{hyperref}       % hyperlinks
\usepackage{url}            % simple URL typesetting
\usepackage{booktabs}       % professional-quality tables
\usepackage{amsfonts}       % blackboard math symbols
\usepackage{nicefrac}       % compact symbols for 1/2, etc.
\usepackage{microtype}      % microtypography
\usepackage{xcolor}         % colors

\usepackage{amsmath, amsfonts}
\usepackage{amssymb}
\usepackage{stmaryrd}
\usepackage{graphicx}
\usepackage{multirow}
\usepackage{subcaption}
\usepackage{tikz}
\usepackage{pgfplots}
\usepackage{mathtools}
\usepackage{svg}

\newcommand{\x}{\mathbf{x}}
\newcommand{\s}{\boldsymbol{s}}

\newcommand{\W}{\mathbf{W}}
\newcommand{\f}{\boldsymbol{f}}
\newcommand{\rd}{\mathrm{d}}
\newcommand{\bmu}{\boldsymbol{\mu}}
\newcommand{\ftheta}{\boldsymbol{f_\theta}}
\newcommand{\fsolver}{\boldsymbol{f_\texttt{solver}}}

\definecolor{blue}{HTML}{007FFF}
\definecolor{orange}{HTML}{FF8000}

\title{Inconsistencies In Consistency Models: Better ODE Solving Does Not Imply Better Samples}

\author{%
  No\"el Vouitsis\\
  Layer 6 AI\\
  \And
   Rasa Hosseinzadeh\\
   Layer 6 AI\\
   \And
   Brendan Leigh Ross\\
   Layer 6 AI\\
   \And
   Valentin Villecroze\\
   Layer 6 AI\\
   \And
   Satya Krishna Gorti\\
   Layer 6 AI\\
   \And
   Jesse C. Cresswell\\
   Layer 6 AI\\
   \And
   Gabriel Loaiza-Ganem \\
   Layer 6 AI\\
   \AND
   \texttt{\{noel, rasa, brendan, valentin.v, satya, jesse, gabriel\}@layer6.ai}
}

\begin{document}

\maketitle

\begin{abstract}
  Although diffusion models can generate remarkably high-quality samples, they are intrinsically bottlenecked by their expensive iterative sampling procedure. Consistency models (CMs) have recently emerged as a promising diffusion model distillation method, reducing the cost of sampling by generating high-fidelity samples in just a few iterations. Consistency model distillation aims to solve the probability flow ordinary differential equation (ODE) defined by an existing diffusion model. CMs are not directly trained to minimize error against an ODE solver, rather they use a more computationally tractable objective. As a way to study how effectively CMs solve the probability flow ODE, and the effect that any induced error has on the quality of generated samples, we introduce Direct CMs, which \textit{directly} minimize this error. Intriguingly, we find that Direct CMs reduce the ODE solving error compared to CMs but also result in significantly worse sample quality, calling into question why exactly CMs work well in the first place. Full code is available at: \url{https://github.com/layer6ai-labs/direct-cms}.
\end{abstract}

\section{Introduction}

In recent years, diffusion models (DMs) \cite{sohl2015deep, ho2020denoising} have become the de facto standard generative models \citep{loaiza2024deep} for many perceptual data modalities such as images \cite{ramesh2022hierarchical, saharia2022photorealistic, rombach2022high, dai2023emu}, video \cite{ho2022video, blattmann2023stable, wu2023tune, wang2024videocomposer}, and audio \cite{kong2020diffwave, ruan2023mm, huang2023make}. Despite their successes, an inherent drawback of diffusion models stems from their iterative sampling procedure, whereby hundreds or thousands of function calls to the diffusion model are typically required to generate high-quality samples, limiting their practicality in low-latency settings. A prominent approach for improving the sampling efficiency of diffusion models is to subsequently distill them into models capable of few-step generation \cite{luhman2021knowledge, salimans2022progressive, poole2022dreamfusion, meng2023distillation, berthelot2023tract, gu2023boot, zheng2023fast, yin2024one, sauer2023adversarial}. Among the works in this vein, consistency models (CMs) \cite{song2023consistency} have garnered attention due to their simple premise as well as their ability to successfully generate samples with only a few steps. CMs leverage the ordinary differential equation (ODE) formulation of diffusion models, called the probability flow (PF) ODE, that defines a deterministic mapping between noise and data \cite{song2020score}. The goal of consistency model distillation is to train a model (the student) to solve the PF ODE of an existing diffusion model (the teacher) from all points along any ODE trajectory in a single step. 
The loss proposed by \citet{song2023consistency} to train CMs does not directly minimize the error against an ODE solver; the solver is mimicked only at optimality and under the assumptions of arbitrarily flexible networks and perfect optimization. We thus hypothesize that the error against the ODE solver can be further driven down by \textit{directly} solving the PF ODE at each step using strong supervision from the teacher, which we call a direct consistency model (Direct CM). Although Direct CMs are more expensive to train than standard CMs, they provide a relevant tool to probe how well CMs solve the PF ODE and how deviations from an ODE solver affect sample quality. We perform controlled experiments to compare CMs and Direct CMs using a state-of-the-art and large-scale diffusion model from the Stable Diffusion family \cite{rombach2022high}, SDXL \cite{podell2023sdxl}, as the teacher model for distillation. We show that Direct CMs perform better at solving the PF ODE but, surprisingly, that they translate to noticeably worse sample quality. This unexpected result challenges the conception that better ODE solving necessarily implies better sample quality, a notion that is implicitly assumed by CMs and its variations alike \cite{song2020improved, geng2024consistency, heek2024multistep, zheng2024trajectory, li2024bidirectional}. Our findings serve as a counterexample to this statement, thus calling into question the community's understanding of ODE-based diffusion model distillation and its implications on sample quality. Since CMs achieve larger ODE solving error, we surmise that other confounding factors contribute to their improved sample quality. We thus call for additional investigation to clarify this seemingly paradoxical behaviour of ODE-based diffusion model distillation.

\section{Background and Related Work}\label{sec:back}

\paragraph{Diffusion Models} The overarching objective of diffusion models is to learn to reverse a noising process that iteratively transforms data into noise. In the limit of infinite noising steps, this iterative process can be formalized as a stochastic different equation (SDE), called the \textit{forward SDE}. The goal of diffusion models amounts to reversing the forward SDE,  hence mapping noise to data \cite{song2020score}. 

Formally, denoting the data distribution as $p_0$, the forward SDE is given by
\begin{equation}
\label{eq:forward_sde}
    \rd \x_t = \bmu(\x_t, t) \rd t + \sigma(t) \rd \W_t
    ,\quad \x_0 \sim p_0,
\end{equation}
where $t \in [0, T]$ for some fixed $T$, $\bmu$ and $\sigma$ are hyperparameters, and $\W_t$ denotes a multivariate Brownian motion. We denote the implied marginal distribution of $\x_t$ as $p_t$; the intuition here is that, with correct choice of hyperparameters, $p_T$ is almost pure noise. \citet{song2020score} showed that the following ODE, referred to as the probability flow (PF) ODE, shares the same marginals as the forward SDE,
\begin{equation}
\label{eq:forward_ode}
    \rd \x_t = \left(\bmu(\x_t, t) - \dfrac{\sigma^2(t)}{2}\nabla \log p_t(\x_t)\right)\rd t,
\end{equation}
where $\nabla \log p_t$ is the (Stein) score function. In other words, if the PF ODE is started at $\x_0 \sim p_0$, then $\x_t \sim p_t$. Under standard regularity conditions, for any initial condition $\x_0$ this ODE admits a unique trajectory $(\x_t)_{t \in [0,T]}$ as a solution. Thus, any point $\x_t$ uniquely determines the entire trajectory, meaning that \autoref{eq:forward_ode} implicitly defines a deterministic mapping $\f_*:(\x_{t}, t, t') \mapsto \x_{t'}$ which can be computed by solving \autoref{eq:forward_ode} backward through time whenever $t>t'$. In principle this function can be used to sample from $p_0$, since $\f_*(\x_T, T, 0)$ will be distributed according to $p_0$ if $\x_T \sim p_T$. In practice this cannot be done exactly, and three approximations are performed. First, the score function is unknown, and diffusion models train a neural network $\s(\x_t, t)$ to approximate it, i.e., \ $\s(\x_t, t) \approx \nabla \log p_t(\x_t)$. This approximation results in the new PF ODE, sometimes called the empirical PF ODE,
\begin{equation}
\label{eq:forward_emp_ode}
    \rd \x_t = \left(\bmu(\x_t, t) - \dfrac{\sigma^2(t)}{2}\s(\x_t, t)\right)\rd t,
\end{equation}
whose solution function we denote as $\f_s$. Second, computing $\f_s(\x_T, T, 0)$ still requires solving an ODE, meaning that a numerical solver must be used to approximate it. We denote the solution of a numerical ODE solver as $\fsolver$, and a single step of the solver from time $t$ to time $t'$ as $\Phi(\cdot, t, t')$. More formally, discretizing the interval $[0,T]$ as $0=t_0 < \dots < t_N = T,$ we have that whenever $n>m$, $\fsolver(\x_{t_n}, t_n, t_m)$ is defined recursively as $\fsolver(\x_{t_n}, t_n, t_m) = \widehat{\x}_{t_m}$ where $\widehat{\x}_{t_{i-1}} = \Phi(\widehat{\x}_{t_i}, t_i, t_{i-1})$ for $i=n, n-1, \dots, m+1$ with $\widehat{\x}_{t_n}=\x_{t_n}$.
Lastly, $p_T$ is also unknown, but since it is very close to pure noise, it can be approximated with an appropriate Gaussian distribution $\widehat{p}_T$.

In summary, by leveraging the empirical PF ODE, samples from a diffusion model can be obtained as $\fsolver(\x_T, T, 0),$ where $\x_T \sim \widehat{p}_T$. If the approximations made throughout are accurate, then $\fsolver \approx \f_s \approx \f_*$ and $\widehat{p}_T \approx p_T$, so that samples from the model resemble samples from $p_0$ \citep{chen2023sampling}. Despite their ability to generate high-quality samples, an inherent drawback of DMs is rooted in their sampling procedure, since computing $\Phi$ requires a function call to $\s$; the iterative refinement of denoised samples to generate high-quality solution trajectories is computationally intensive.

\paragraph{Consistency Models} Consistency models \cite{song2023consistency} leverage the PF ODE formulation of DMs to enable few-step generation. They can be used either for DM distillation or trained standalone from scratch; we only consider distillation in our work since the score function of pre-trained DMs gives us a tool to directly study the effect of ODE solving on CMs. Given a trained diffusion model $\s$ with a corresponding $\fsolver$, the idea of consistency model distillation is to train a neural network $\ftheta$ such that $\ftheta(\x_{t_n}, t_n) \approx \fsolver(\x_{t_n}, t_n, 0)$ for all $n \in \{1,\dots,N\}$. In other words, CMs aim to learn a function to mimic the solver of the empirical PF ODE, thus circumventing the need to repeatedly evaluate $\s$ during sampling. CMs learn $\ftheta$ by enforcing the self-consistency property, meaning that for every $\x_{t_n}$ and $\x_{t_{n'}}$ along the same trajectory, $\ftheta(\x_{t_n},t_n)$ and $\ftheta(\x_{t_{n'}}, t_{n'})$ are encouraged to match. More specifically, CMs are trained by minimizing the consistency distillation loss,
\begin{equation}
\label{eq:cm_loss}
    \mathcal{L}_\text{CD} := \mathbb{E}_{\x_0 \sim p_0,n \sim \mathcal{U} \llbracket 1,N \rrbracket , \x_{t_n} \sim p_{t_n \mid 0}(\cdot \mid \x_0)} \Big[\lambda(t_n) d\Big( \ftheta(\x_{t_n}, t_n), \boldsymbol{f_{\bar{\theta}}}(\widehat{\x}_{t_{n-1}}, t_{n-1})\Big)\Big],
\end{equation}
where $p_{t|0}$ is the transition kernel corresponding to \autoref{eq:forward_sde}, $\lambda > 0$ is a weighting function treated as a hyperparameter, $d$ is any distance, $\boldsymbol{\bar{\theta}}$ is a frozen version of $\boldsymbol{\theta}$, and $\widehat{\x}_{t_{n-1}} = \Phi(\x_{t_n}, t_n, t_{n-1})$. Since the transition kernel is given by a known Gaussian, the above objective is tractable. CMs parameterize $\ftheta$ in such a way that $\ftheta(\widehat{\x}_0, 0) = \widehat{\x}_0$ holds. This property is referred to as the boundary condition, and prevents \autoref{eq:cm_loss} from being pathologically minimized by $\ftheta$ collapsing onto a constant function.

During sampling, CMs can use one or multiple function evaluations of $\ftheta$, enabling a trade-off between computational cost and sample quality. For example, if given a budget of two function evaluations, rather than produce a sample as $\ftheta(\x_T, T)$, one could run \autoref{eq:forward_sde} until some time $t_{n'}$ starting from $\ftheta(\x_T, T)$ to produce $\x_{t_{n'}}$, and then output $\ftheta(\x_{t_{n'}}, t_{n'})$ as the sample. This idea generalizes to more function evaluations, although note that $\ftheta(\x_{t_{n'}}, t_{n'})$ and $\x_{t_{n'}}$ do not belong to the same ODE trajectory as $\ftheta(\x_T, T)$ and $\x_T$ due to the added noise from the forward SDE.

\section{Direct Consistency Models}

\begin{figure}[t]
   \centering
   \hspace*{-1cm}
   \includegraphics{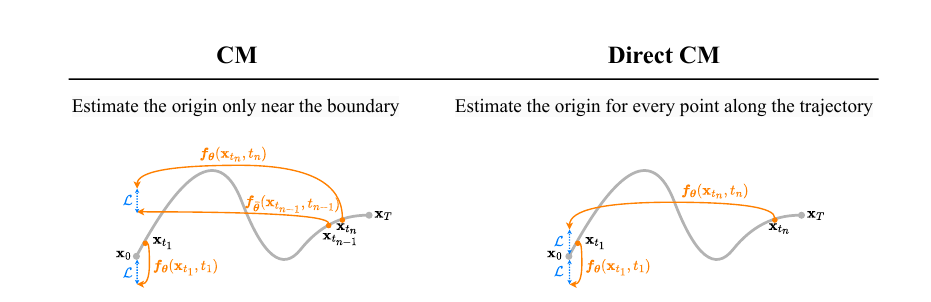}
 \caption{CMs (left) are weakly supervised ODE solvers, only learning to map points along a trajectory that are near the trajectory's origin back to the origin itself; points that are far from the origin instead enforce a self-consistency property, relying on weak self-supervision to solve the PF ODE. Direct CMs (right) are strongly supervised ODE solvers, instead learning to directly map all points along a trajectory back to the origin.}
 \label{fig:diag}
\end{figure}

In \autoref{eq:cm_loss}, $\x_0$ and $\x_{t_n}$ do not belong to the same ODE trajectory since noise is added to obtain $\x_{t_n}$ from $\x_0$ via the forward SDE's transition kernel. Thus, it would not make sense to enforce consistency by minimizing $d(\ftheta(\x_{t_n}, t_n), \x_0)$, and \autoref{eq:cm_loss} is used instead. 
While \citet{song2023consistency} theoretically showed that perfectly minimizing \autoref{eq:cm_loss} with an arbitrarily flexible $\ftheta$ results in $\ftheta(\x_{t_n}, t_n) = \fsolver(\x_{t_n}, t_n, 0)$, 
in practice it has been observed that CMs can be difficult to optimize, with slow convergence or, in some cases, divergence \cite{heek2024multistep, geng2024consistency}. 
We attribute this behaviour to what we call ``weak supervision'' in the CM loss, namely that $\ftheta$ is not directly trained to map $\x_{t_n}$ to the origin of its ODE trajectory. 
The constraint that the CM should map any point on the ODE trajectory to the trajectory's origin is only weakly enforced through the boundary condition parameterization of $\ftheta$. Only at time $t_1$ does the objective directly encourage mapping points $\x_{t_1}$ to the trajectory's origin. The network $\ftheta$ must therefore first learn to map slightly noised data back to the origin before that constraint can be properly enforced for noisier inputs at larger timesteps. 
We depict this behaviour in \autoref{fig:diag} (left).

In order to assess the impact of ODE solving on CMs, we put forth a more intuitive and interpretable variation of its loss as
\begin{equation}
\label{eq:dir_cm_loss}
    \mathcal{L}_\text{CD}^{\text{Direct}} := \mathbb{E}_{\x_0 \sim p_0,n \sim \mathcal{U} \llbracket 1,N \rrbracket , \x_{t_n} \sim p_{t_n \mid 0}(\cdot \mid \x_0)} \Big[\lambda(t_n) d\Big( \ftheta(\x_{t_n}, t_n), \fsolver(\x_{t_n}, t_n, 0)  \Big)\Big],
\end{equation}
where we \textit{directly} enforce that all points along a trajectory map to its origin, rather than providing only weak supervision as in CMs; see \autoref{fig:diag} (right). 
We see this loss as the smallest possible modification to CMs resulting in the direct matching of the model and the solver. Note that unlike standard CMs, Direct CMs do not require enforcing the boundary condition in the parameterization of $\ftheta$ to prevent collapse, although it is of course still valid to do so. 
While this loss requires solving the ODE for $n$ steps at each iteration and is therefore more computationally expensive than \autoref{eq:cm_loss}, we only propose this formulation for comparative purposes rather than suggesting its use in practice. 

As we will show, \autoref{eq:dir_cm_loss} does indeed solve the empirical PF ODE better than \autoref{eq:cm_loss} but, intriguingly, it translates to worse sample quality. We define the ODE solving error $\mathcal{E}$ as the expected distance between the ODE solver's solution and the CM's prediction with the same initial noise, i.e.,\
\begin{equation}
\label{eq:ode_error}
    \mathcal{E} := \mathbb{E}_{\x_{T} \sim \widehat{p}_T} \Big[d\Big( \boldsymbol{f_\theta}(\mathbf{x}_{T}, T), \fsolver(\mathbf{x}_{T}, T, 0) \Big)\Big].
\end{equation}

\section{Experiments}
\begin{table*}[t]
    \centering
    \renewcommand{\arraystretch}{1.1}
    \caption{Results of ODE solving and image quality for single-step generation. CMs perform worse at solving the PF ODE but produce higher quality images.}
    \begin{tabular}{c|c|c|cccc}
         \toprule
         \multirow{2}{*}{$\Phi$} & \multirow{2}{*}{Method} & ODE & \multicolumn{4}{c}{Image} \\
         \cline{3-7}
         & & $\mathcal{E} \downarrow$ & FID $\downarrow$ & FD-DINO $\downarrow$ & CLIP $\uparrow$ & Aes $\uparrow$ \\
         \hline & \\[-2.1ex]
         \hline
         \multirow{2}{*}{DDIM} & CM & 0.29 & \textbf{\color{blue}103.9} & \textbf{\color{blue}816.3} & \textbf{\color{blue}0.21} & \textbf{\color{blue}5.6} \\
         & Direct CM & \textbf{\color{blue}0.25} & 158.6 & 1095 & 0.20 & 5.1 \\
         \hline
         \multirow{2}{*}{Euler} & CM & 0.29 & \textbf{\color{blue}95.3} & \textbf{\color{blue}747.7} & \textbf{\color{blue}0.21} & \textbf{\color{blue}5.5} \\
         & Direct CM & \textbf{\color{blue}0.23} & 166.0 & 1148 & 0.19 & 5.0 \\
         \hline
         \multirow{2}{*}{Heun} & CM & 0.30 & \textbf{\color{blue}120.5} & \textbf{\color{blue}846.1} & \textbf{\color{blue}0.21} & \textbf{\color{blue}5.5} \\
         & Direct CM & \textbf{\color{blue}0.25} & 162.0 & 1126 & 0.19 & 5.1 \\
         \bottomrule   
    \end{tabular}
    \label{tab:results}
\end{table*}
\paragraph{Training} For all of our experiments, we aim to compare CMs and Direct CMs using large-scale and state-of-the-art DMs trained on Internet-scale data to better reflect the performance of these models in practical real-world settings. Hence, we select SDXL \cite{podell2023sdxl} as the DM to distill, a text-to-image latent diffusion model \cite{rombach2022high} with a 2.6 B parameter U-Net backbone \cite{ronneberger2015u}, capable of generating images at a 1024 px resolution. Classifier-free guidance \cite{ho2022classifier} is commonly used to improve sample quality in text-conditional DMs, so we augment $\s$ in \autoref{eq:forward_emp_ode} as $\s(\x_t, t, c, \omega)$, where $c$ is the text prompt and $\omega$ is the guidance scale, following \citet{luo2023latent}. When distilling a DM, it is common to initialize the student network from the weights of the teacher network so that, in effect, distillation is reduced to a fine-tuning task which requires much less data and resources. We further leverage modern best practices for efficient fine-tuning using low-rank adapters \cite{hu2021lora, luo2023lcm}. We use a high-quality subset of the LAION-5B dataset \cite{schuhmann2022laion} called LAION-Aesthetics-6.5+ for training similar to \citet{luo2023latent}. To ensure a controlled comparison of CMs and Direct CMs, the only component in the code that we modify is the loss. See \autoref{app:app-hyper} for a list of training hyperparameters. 
\paragraph{Evaluation}
We perform quantitative comparisons using metrics that measure ODE solving quality as well as image quality. For ODE solving, we use $\mathcal{E}$ (\autoref{eq:ode_error}, lower is better) which is only valid for single-step generation.\footnote{As mentioned in \autoref{sec:back}, multi-step sampling in CMs requires adding random noise to the model's prediction using the forward SDE. However, the noised prediction will map to a different underlying PF ODE trajectory, so comparing it to the original trajectory would not give a meaningful metric for ODE-solving fidelity.} For image metrics, we use Fréchet Distance on Inception (FID \cite{heusel2017gans}, lower is better) and DINOv2 (FD-DINO \cite{oquab2023dinov2, stein2024exposing}, lower is better) latent spaces to assess distributional quality, CLIP score (CLIP \cite{radford2021learning, hessel2021clipscore}, higher is better) for prompt-image alignment, and aesthetic score (Aes \cite{ren2024hyper}, higher is better) as a proxy to subjective visual appeal. All generated samples use fixed seeds to ensure consistent random noise. The reference dataset for both FID and FD-DINO uses 10k samples generated from the teacher with the same seeds.

\begin{figure}[t!]
   \centering
   \hspace*{-1cm}
   \includegraphics[width=\textwidth]{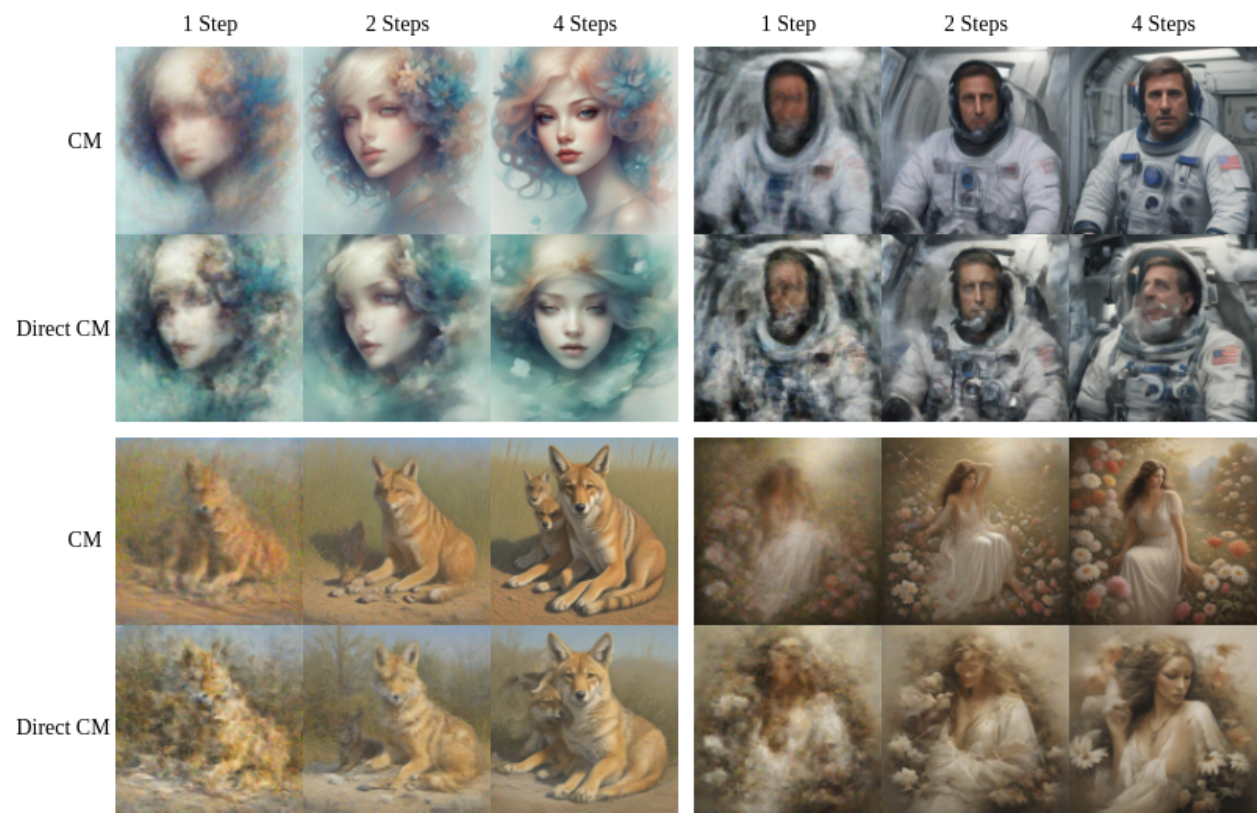}
 \caption{Samples generated by both CMs and Direct CMs. The samples produced by CMs are clearly of higher quality. All corresponding images are generated from the same initial noise.}
 \label{fig:qual}
\end{figure}

\paragraph{Quantitative Analysis} We provide a quantitative evaluation of CMs and Direct CMs in \autoref{tab:results}. We show performance for three different choices of numerical ODE solvers $\Phi$, namely DDIM \cite{song2020denoising} following \citet{luo2023latent}, Euler \cite{euler2010griffiths}, and Heun \cite{heun1995heun} following \citet{song2023consistency}. As mentioned earlier, $\mathcal{E}$ is a meaningful metric for ODE-solving fidelity only for single-step generation, so we focus our main quantitative analysis on single-step generation; we provide additional image-based metrics for two- and four-step generation in \autoref{app:app-quant} for completeness. Across all image-based metrics in \autoref{tab:results}, we observe that CMs convincingly outperform Direct CMs, meaning that training with \autoref{eq:cm_loss} results in largely superior image quality than training with \autoref{eq:dir_cm_loss}. However, in terms of their ability to more accurately solve the PF ODE, we find that Direct CMs are consistently better. Ironically, the objective of CMs, as presented by \citet{song2023consistency}, is motivated by learning to faithfully solve the PF ODE, so it is highly surprising that more accurate solving can translate to worse image quality.

Our experiments suggest that the pursuit of diffusion model distillation methods to better solve the PF ODE might be a red herring, and that it is not in complete alignment with the goal of generating high-quality samples. 
Several follow-up works to CMs \cite{song2023improved, geng2024consistency} have further built upon the PF ODE formulation, proposing variations to CMs such as splitting the trajectory into segments \cite{heek2024multistep, zheng2024trajectory} or learning to solve the ODE bidirectionally \cite{li2024bidirectional} for example. Although they observed better sample quality, we reject the notion that their improvements are strictly entailed by better PF ODE solving. Our results in \autoref{tab:results} suggest that the high quality of images produced by ODE solving methods (such as CMs and variations) cannot be fully attributed to their ODE solving fidelity; confounding factors should be considered as well.

Moreover, we argue that this observed discrepancy between ODE solving and sample quality might suggest that PF ODE solving on its own may not be the most reliable approach to distill a diffusion model in practice. It is perhaps unsurprising then that several follow-up works improving upon CMs rely on auxiliary losses to supplement ODE solving such as adversarial \cite{kim2023consistency, chadebec2024flash, wang2024phased, kim2024pagoda}, distribution matching \cite{chadebec2024flash, ren2024hyper}, and human feedback learning \cite{xie2024mlcm, ren2024hyper} losses.

\paragraph{Qualitative Comparison} We corroborate our observations with a qualitative comparison of CMs and Direct CMs in \autoref{fig:qual}, and include additional samples in Appendix \ref{app:app-qual}. We show generated samples from a fixed seed for one, two and four sampling steps using text prompts from the training set. It is clear that CMs produce higher-quality samples than Direct CMs with better high frequency details and fewer artifacts. 

\begin{figure}[tbp]
    \centering
    \scriptsize
    \begin{subfigure}[t]{0.19\textwidth}
        \centering
        \begin{tikzpicture}
              \begin{axis}[
                width=3.2cm,
                height=3.2cm,
                xlabel={$\omega$},
                ylabel={$\mathcal{E}$ $\downarrow$},
                xlabel style={yshift=0.2cm},
                ylabel style={yshift=-0.45cm},
                grid=major,
                xtick={0, 5, 10},
                ytick={0.10, 0.20, 0.30},
                legend style={at={(3.7, 1.4)}, draw=none, anchor=north, legend columns=-1}
                ]
                % Your data and plot commands go here
                \addplot[blue, mark=*, mark size=1, style=thick] coordinates {(1,0.08977) (4,0.1733) (8,0.269) (11,0.3315)};
                \addplot[orange, mark=*, mark size=1, style=thick] coordinates {(1,0.07064) (4,0.141) (8,0.228) (11,0.2966)};
                \legend{CM, Direct CM}
              \end{axis}
        \end{tikzpicture}
        \label{fig:w-mse}
    \end{subfigure}
    \begin{subfigure}[t]{0.19\textwidth}
        \centering
        \begin{tikzpicture}
              \begin{axis}[
                width=3.2cm,
                height=3.2cm,
                xlabel={$\omega$},
                ylabel={FID $\downarrow$},
                xlabel style={yshift=0.2cm},
                ylabel style={yshift=-0.55cm},
                grid=major,
                xtick={0, 5, 10},
                ytick={100, 150},
                ]
                % Your data and plot commands go here
                \addplot[blue, mark=*, mark size=1, style=thick] coordinates {(1,72.06) (4,94.63) (8,118.3) (11,161.5)};
                \addplot[orange, mark=*, mark size=1, style=thick] coordinates {(1,107.7) (4,138) (8,154.7) (11,168.3)};
              \end{axis}
        \end{tikzpicture}
        \label{fig:w-fid}
    \end{subfigure}
    \begin{subfigure}[t]{0.19\textwidth}
        \centering
        \begin{tikzpicture}
              \begin{axis}[
                width=3.2cm,
                height=3.2cm,
                xlabel={$\omega$},
                ylabel={FD-DINO $\downarrow$},
                xlabel style={yshift=0.2cm},
                ylabel style={yshift=-0.35cm},
                grid=major,
                xtick={0, 5, 10},
                ytick={500, 1000},
                ]
                % Your data and plot commands go here
                \addplot[blue, mark=*, mark size=1, style=thick] coordinates {(1,528.9) (4,692.6) (8,844.7) (11,1091)};
                \addplot[orange, mark=*, mark size=1, style=thick] coordinates {(1,816.7) (4,1031) (8,1083) (11,1153)};
              \end{axis}
        \end{tikzpicture}
        \label{fig:w-fdd}
    \end{subfigure}
    \begin{subfigure}[t]{0.22\textwidth}
        \centering
        \begin{tikzpicture}
              \begin{axis}[
                width=3.2cm,
                height=3.2cm,
                xlabel={$\omega$},
                ylabel={CLIP $\uparrow$},
                xlabel style={yshift=0.2cm},
                ylabel style={yshift=-0.45cm},
                grid=major,
                xtick={0, 5, 10},
                ytick={0.2, 0.21, 0.22},
                ]
                % Your data and plot commands go here
                \addplot[blue, mark=*, mark size=1, style=thick] coordinates {(1,0.2123) (4,0.2076) (8,0.2044) (11,0.1944)};
                \addplot[orange, mark=*, mark size=1, style=thick] coordinates {(1,0.1945) (4,0.1972) (8,0.197) (11,0.1918)};
              \end{axis}
        \end{tikzpicture}
        \label{fig:w-clip}
    \end{subfigure}
    \begin{subfigure}[t]{0.19\textwidth}
        \centering
        \hspace{-5ex}
        \begin{tikzpicture}
              \begin{axis}[
                width=3.2cm,
                height=3.2cm,
                xlabel={$\omega$},
                ylabel={Aes $\uparrow$},
                xlabel style={yshift=0.2cm},
                ylabel style={yshift=-0.6cm},
                grid=major,
                xtick={0, 5, 10},
                ytick={5.0, 5.4, 5.8},
                ]
                % Your data and plot commands go here
                \addplot[blue, mark=*, mark size=1, style=thick] coordinates {(1,5.792) (4,5.546) (8,5.402) (11,5.192)};
                \addplot[orange, mark=*, mark size=1, style=thick] coordinates {(1,5.549) (4,5.432) (8,5.279) (11,5.098)};
              \end{axis}
        \end{tikzpicture}
        \label{fig:w-aes}
    \end{subfigure}
    \vspace{-3ex}
    \caption{Effect of the teacher's guidance scale $\omega$. We use $N=50$ here for faster experimentation. In all cases, we observe that Direct CMs are better at solving the PF ODE, but CMs produce higher quality images.}
    \label{fig:w-ablate}
\end{figure}
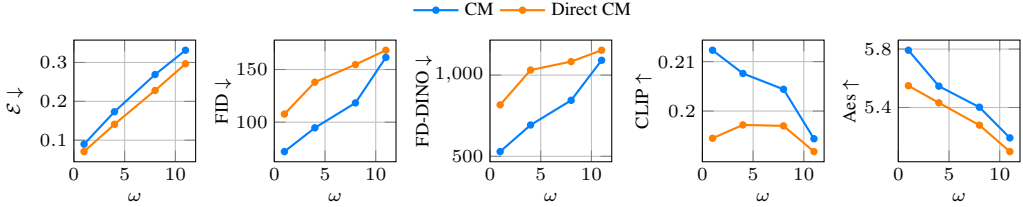

\paragraph{Ablations} To ensure that our findings are agnostic to hyperparameter selection in the underlying PF ODE and ODE solver, we sweep over various discretization intervals $N \in \{25, 50, 100, 200\}$ and guidance scales $\omega \in \{1, 4, 8, 11\}$, and provide results for single-step generation in \autoref{fig:n-ablate} and \autoref{fig:w-ablate}. Regardless of the teacher's guidance scale and discretization, Direct CMs solve the PF ODE more accurately, yet CMs produce higher quality images.

\begin{figure}[tbp]
    \centering
    \scriptsize
    \begin{subfigure}[t]{0.19\textwidth}
        \centering
        \begin{tikzpicture}
              \begin{axis}[
                width=3.2cm,
                height=3.2cm,
                xlabel={$N$},
                ylabel={$\mathcal{E}$ $\downarrow$},
                xlabel style={yshift=0.2cm},
                ylabel style={yshift=-0.45cm},
                grid=major,
                xtick={0, 100, 200},
                ytick={0.20, 0.25, 0.30},
                legend style={at={(3.7, 1.4)}, draw=none, anchor=north, legend columns=-1}
                ]
                % Your data and plot commands go here
                \addplot[blue, mark=*, mark size=1, style=thick] coordinates {(25,0.2376) (50,0.269) (100,0.2925) (200,0.3339)};
                \addplot[orange, mark=*, mark size=1, style=thick] coordinates {(25,0.207) (50,0.228) (100,0.2527) (200,0.2522)};
                \legend{CM, Direct CM}
              \end{axis}
        \end{tikzpicture}
        \label{fig:n-mse}
    \end{subfigure}
    \begin{subfigure}[t]{0.19\textwidth}
        \centering
        \begin{tikzpicture}
              \begin{axis}[
                width=3.2cm,
                height=3.2cm,
                xlabel={$N$},
                ylabel={FID $\downarrow$},
                xlabel style={yshift=0.2cm},
                ylabel style={yshift=-0.55cm},
                grid=major,
                xtick={0, 100, 200},
                ytick={100, 150},
                ]
                % Your data and plot commands go here
                \addplot[blue, mark=*, mark size=1, style=thick] coordinates {(25,97.4) (50,118.3) (100,103.9) (200,105.6)};
                \addplot[orange, mark=*, mark size=1, style=thick] coordinates {(25,140.2) (50,154.7) (100,158.6) (200,167.1)};
              \end{axis}
        \end{tikzpicture}
        \label{fig:n-fid}
    \end{subfigure}
    \begin{subfigure}[t]{0.19\textwidth}
        \centering
        \begin{tikzpicture}
              \begin{axis}[
                width=3.2cm,
                height=3.2cm,
                xlabel={$N$},
                ylabel={FD-DINO $\downarrow$},
                xlabel style={yshift=0.2cm},
                ylabel style={yshift=-0.35cm},
                grid=major,
                xtick={0, 100, 200},
                ytick={800, 1000},
                ]
                % Your data and plot commands go here
                \addplot[blue, mark=*, mark size=1, style=thick] coordinates {(25,769.1) (50,844.7) (100,816.3) (200,799.5)};
                \addplot[orange, mark=*, mark size=1, style=thick] coordinates {(25,1024) (50,1083) (100,1095) (200,1151)};
              \end{axis}
        \end{tikzpicture}
        \label{fig:n-fdd}
    \end{subfigure}
    \begin{subfigure}[t]{0.22\textwidth}
        \centering
        \begin{tikzpicture}
              \begin{axis}[
                width=3.2cm,
                height=3.2cm,
                xlabel={$N$},
                ylabel={CLIP $\uparrow$},
                xlabel style={yshift=0.2cm},
                ylabel style={yshift=-0.45cm},
                grid=major,
                xtick={0, 100, 200},
                ytick={0.2, 0.21, 0.22},
                ]
                % Your data and plot commands go here
                \addplot[blue, mark=*, mark size=1, style=thick] coordinates {(25,0.215) (50,0.2044) (100,0.2105) (200,0.2078)};
                \addplot[orange, mark=*, mark size=1, style=thick] coordinates {(25,0.2054) (50,0.197) (100,0.1963) (200,0.1922)};
              \end{axis}
        \end{tikzpicture}
        \label{fig:n-clip}
    \end{subfigure}
    \begin{subfigure}[t]{0.19\textwidth}
        \centering
        \hspace{-5ex}
        \begin{tikzpicture}
              \begin{axis}[
                width=3.2cm,
                height=3.2cm,
                xlabel={$N$},
                ylabel={Aes $\uparrow$},
                xlabel style={yshift=0.2cm},
                ylabel style={yshift=-0.6cm},
                grid=major,
                xtick={0, 100, 200},
                ytick={5.2, 5.6},
                ymax=5.8,
                ]
                % Your data and plot commands go here
                \addplot[blue, mark=*, mark size=1, style=thick] coordinates {(25,5.517) (50,5.402) (100,5.556) (200,5.434)};
                \addplot[orange, mark=*, mark size=1, style=thick] coordinates {(25,5.458) (50,5.279) (100,5.148) (200,5.145)};
              \end{axis}
        \end{tikzpicture}
        \label{fig:n-aes}
    \end{subfigure}
    \vspace{-3ex}
    \caption{Effect of the teacher's number of discretization intervals $N$. In all cases, we observe that Direct CMs are better at solving the PF ODE, but CMs produce higher quality images.}
    \label{fig:n-ablate}
\end{figure}
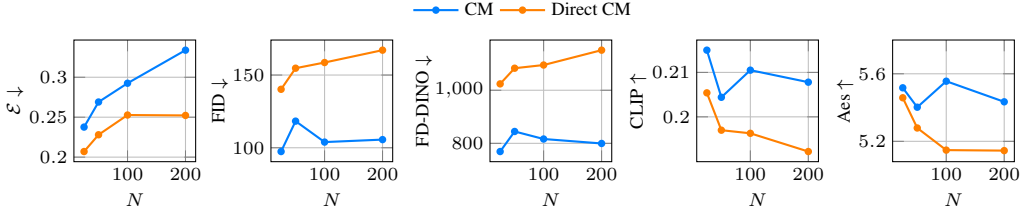

\section{Conclusions and Future Work}
Although consistency models have achieved success in distilling diffusion models into few-step generators, we find that there exists a gap between their theory and practice. Solving the PF ODE is central to the theoretical motivation of CMs, but we show that we can solve the same PF ODE more accurately using Direct CMs while generating samples of noticeably worse quality. Naturally, we question what additional underlying factors might be contributing to the effectiveness of CMs, and call for additional research from the community to bridge this observed gap between solving the PF ODE and generating high-quality samples. We finish by putting forth some potential explanations: $(i)$ since our experiments are carried out with latent diffusion models, the ODEs are defined on the corresponding latent space, and it could be that the closeness to the solver's solutions observed in Direct CMs is undone after decoding to pixel space; $(ii)$ if the pre-trained diffusion model failed to closely approximate the true score function (as could be the case when the true score function is unbounded \cite{pidstrigach2022score, lu2023mathematical, loaiza2024deep}) then $\f_s \not\approx \f_*$, meaning that even if a model closely approximates $\fsolver$ and thus $\f_s$, it need not be the case that it also properly approximates $\f_*$; and $(iii)$ although both the CM and Direct CM objectives (\autoref{eq:cm_loss} and \autoref{eq:dir_cm_loss}, respectively) are meant to mimic the solver $\fsolver$ at optimality, in practice this optimum is never perfectly achieved, and the CM objective might inadvertently provide a beneficial inductive bias which improves sample quality.

\newpage
{
\small
\bibliography{main}
\bibliographystyle{plainnat}
}

%%%%%%%%%%%%%%%%%%%%%%%%%%%%%%%%%%%%%%%%%%%%%%%%%%%%%%%%%%%%
\newpage
\appendix

\section{Appendix}
\subsection{Hyperparameters}
\label{app:app-hyper}

We provide a list of default hyperparameter values in \autoref{tab:hypers}. We only train a small number of LoRA blocks \cite{hu2021lora} following \citet{luo2023lcm}, and find that metric and loss curves stabilized around 250 training steps. 
To enforce the boundary condition in CMs, we follow \citet{song2023consistency} and parameterize $\ftheta(\x_{t_n}, t_n) =  c_{\text{skip}}(t_n) \x_{t_n} + c_{\text{out}}(t_n)F_{\boldsymbol{\theta}}(\x_{t_n}, t_n)$ where $F_{\boldsymbol{\theta}}(\x_{t_n}, t_n)$ in our case is the SDXL U-Net backbone with learnable LoRA blocks, and $c_{\text{skip}}(t_n)$ and $c_{\text{out}}(t_n)$ are differentiable functions such that $c_{\text{skip}}(0)=1$ and $c_{\text{out}}(0)=0$. We set the values of $c_{\text{skip}}(t_n)$ and $c_{\text{out}}(t_n)$ following \citet{luo2023latent} (see \autoref{tab:hypers}), and note that this choice is roughly equivalent to a step function where, for $n \ge 1$, $c_{\text{skip}}(t_n)\approx 0$ and $c_{\text{out}}(t_n)\approx 1$. As mentioned in the main text, Direct CMs do not require enforcing a boundary condition by construction, but we parameterize them identically to CMs in order to ensure controlled experiments so that we can attribute any differences between them solely to differences in the loss. All experiments were performed on a single 48GB NVIDIA RTX 6000 Ada GPU.

\begin{table*}[h]
    \centering
    \renewcommand{\arraystretch}{1.1}
    \caption{Default hyperparameters for both CMs and Direct CMs, unless otherwise specified.}
    \begin{tabular}{l|l}
         \toprule
         Hyperparameter & Default Setting \\
         \hline
         Batch size & 16 \\
         Mixed precision & fp16 \\
         Efficient attention \cite{rabe2021self} & True \\
         Gradient checkpointing & True \\
         Optimizer & 8-bit Adam \cite{dettmers20218} \\
         Adam weight decay & $10^{-2}$ \\
         Num. training steps & 250 \\
         LoRA $r$ \cite{hu2021lora} & 64 \\
         LoRA $\alpha$ \cite{hu2021lora} & 64 \\
         Learning rate scheduler & Constant \\
         Learning rate warmup steps & 0 \\
         Learning rate & $10^{-4}$ \\
         $\Phi$ & DDIM \cite{song2020denoising} \\
         $N$ & 100 \\
         $\omega$ & 8 \\
         $d(\cdot,\cdot)$ & Squared $L_2$ distance \\
         $\lambda(t)$ & 1 \\
         $\sigma_{\text{data}}$ & 0.5 \cite{song2023consistency} \\
         $\tau$ & 10 \\
         $c_{\text{skip}}(t)$ & $\frac{\sigma_{\text{data}}^2}{(t\cdot\tau)^2 + \sigma_{\text{data}}^2}$ \\
         $c_{\text{out}}(t)$ & $\frac{t\cdot\tau}{\sqrt{(t\cdot\tau)^2 + \sigma_{\text{data}}^2}}$ \\
         \bottomrule    
    \end{tabular}
    \label{tab:hypers}
\end{table*}

\subsection{Additional Quantitative Results}
\label{app:app-quant}

\begin{table*}[b!]
    \scriptsize	
    \centering
    \addtolength{\tabcolsep}{-0.4em}
     \renewcommand{\arraystretch}{1.1}
    \caption{Additional image results for two- and four-step generation.}
    \begin{tabular}{c|c|c|ccc|ccc|ccc|ccc}
         \toprule
         \multirow{3}{*}{$\Phi$} & \multirow{3}{*}{Method} & ODE & \multicolumn{12}{c}{Image} \\
         \cline{3-15}
         & & $\mathcal{E} \downarrow$ & \multicolumn{3}{c}{FID $\downarrow$} & \multicolumn{3}{c}{FD-DINO $\downarrow$} & \multicolumn{3}{c}{CLIP $\uparrow$} & \multicolumn{3}{c}{Aes $\uparrow$} \\
         \cline{3-15}
         & & 1-step & 1-step & 2-step & 4-step & 1-step & 2-step & 4-step & 1-step & 2-step & 4-step & 1-step & 2-step & 4-step \\
         \hline & \\[-2.0ex]
         \hline
         \multirow{2}{*}{DDIM} & CM & 0.29 & \textbf{\color{blue}103.9} & \textbf{\color{blue}33.4} & \textbf{\color{blue}19.8}& \textbf{\color{blue}816.3} & \textbf{\color{blue}255.4} & 159.8 & \textbf{\color{blue}0.21} & \textbf{\color{blue}0.27} & 0.27 & \textbf{\color{blue}5.6} & \textbf{\color{blue}6.4} & \textbf{\color{blue}6.7} \\
         & Direct CM & \textbf{\color{blue}0.25} & 158.6 & 55.0 & 21.2 & 1095 & 346.8 & \textbf{\color{blue}155.1} & 0.20 & 0.26 & \textbf{\color{blue}0.28} & 5.1 & 6.2 & 6.5 \\
         \hline
         \multirow{2}{*}{Euler} & CM & 0.29 & \textbf{\color{blue}95.3} & \textbf{\color{blue}27.4} & \textbf{\color{blue}18.9} & \textbf{\color{blue}747.7} & \textbf{\color{blue}221.3} & 156.8 & \textbf{\color{blue}0.21} & \textbf{\color{blue}0.27} & \textbf{\color{blue}0.27} & \textbf{\color{blue}5.5} & \textbf{\color{blue}6.5} & \textbf{\color{blue}6.7} \\
         & Direct CM & \textbf{\color{blue}0.23} & 166.0 & 55.7 & 22.5 & 1148 & 357.3 & \textbf{\color{blue}152.7} & 0.19 & 0.25 & \textbf{\color{blue}0.27} & 5.0 & 6.1 & 6.4 \\
         \hline
         \multirow{2}{*}{Heun} & CM & 0.30 & \textbf{\color{blue}120.5} & \textbf{\color{blue}33.5} & \textbf{\color{blue}20.7} & \textbf{\color{blue}846.1} & \textbf{\color{blue}233.4} & 159.4 & \textbf{\color{blue}0.21} & \textbf{\color{blue}0.27} & 0.27 & \textbf{\color{blue}5.5} & \textbf{\color{blue}6.5} & \textbf{\color{blue}6.7} \\
         & Direct CM & \textbf{\color{blue}0.25} & 162.0 & 54.8 & 21.0 & 1126 & 341.6 & \textbf{\color{blue}150.6} & 0.19 & 0.26 & \textbf{\color{blue}0.28}& 5.1 & 6.2 & 6.4 \\
         \bottomrule    
    \end{tabular}
    \label{tab:results-steps}
\end{table*}

\begin{figure}[htbp]
   \centering
   \hspace*{-1cm}
   \includegraphics[width=\textwidth]{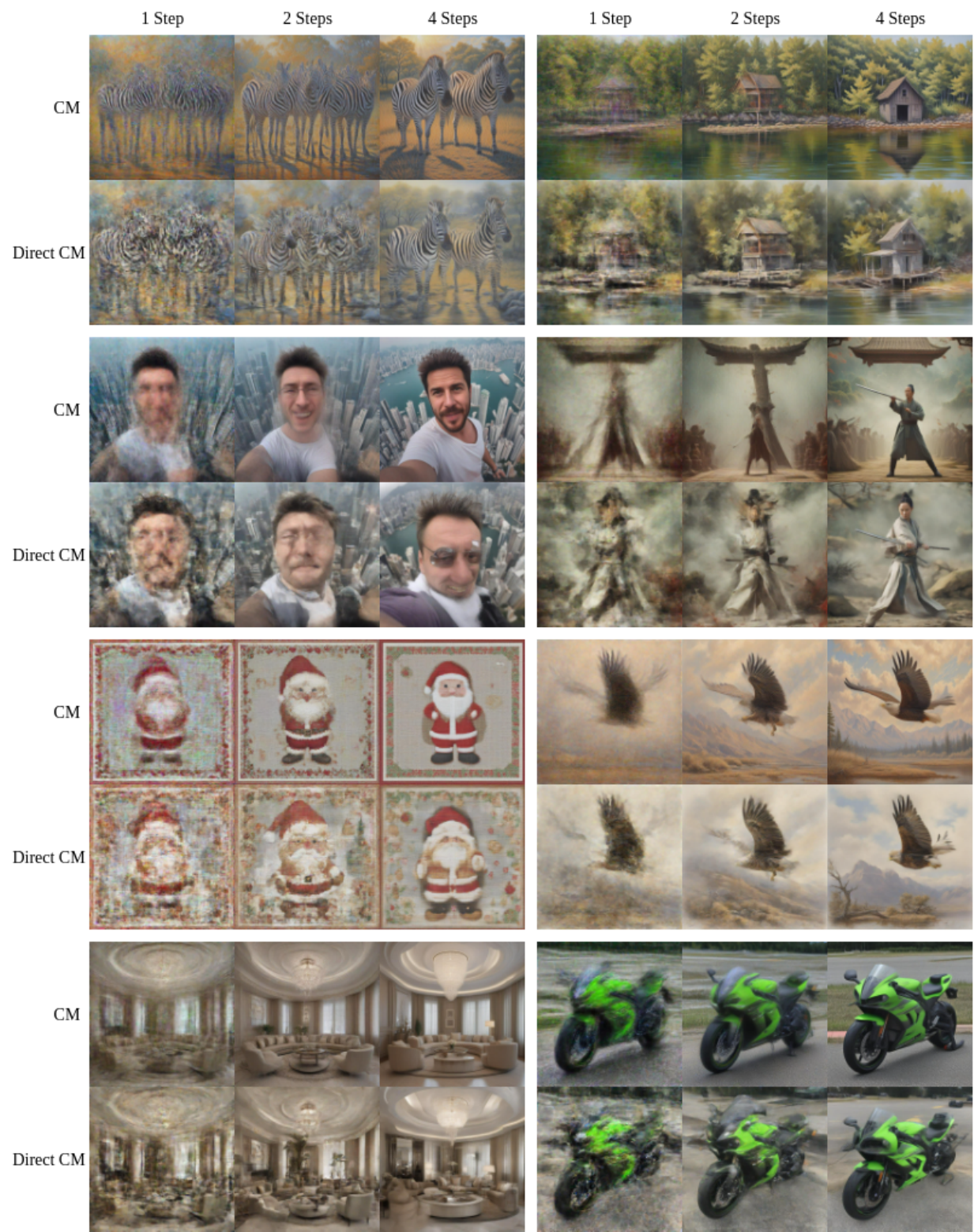}
 \caption{Additional images generated by both CMs and Direct CMs, further highlighting the sample quality difference between the two models. All corresponding images are generated from the same initial noise.}
 \label{fig:qual-app}
\end{figure}

We provide additional quantitative image analysis for two- and four-step generation in \autoref{tab:results-steps}. In almost all cases, these results demonstrate that CMs generate higher quality images than Direct CMs akin to the single-step generation case. Although these results suggest that for four steps Direct CMs slightly outperform CMs in terms of FD-DINO and CLIP score, qualitative comparisons of generated images between both models (see examples in \autoref{fig:qual} and \autoref{fig:qual-app}) quickly reveal that images from CMs have noticeably higher quality. We thus attribute the discrepancy either to imperfections in generative model evaluation metrics as observed by \citet{stein2024exposing}, or to these metrics not perfectly matching aesthetic quality and being affected by additional confounders (e.g., FD-DINO scores are meant to reflect image diversity in addition to image aesthetics). 

\newpage
\subsection{Additional Qualitative Results}
\label{app:app-qual}

\end{document}